\documentclass{article}

\usepackage[final, nonatbib]{nips_2018}

\usepackage[utf8]{inputenc} 
\usepackage[T1]{fontenc}    
\usepackage{hyperref}       
\usepackage{url}            

\usepackage{booktabs}       
\usepackage{amsfonts}       
\usepackage{nicefrac}       
\usepackage{microtype}      

\usepackage[pdftex]{graphicx}
\usepackage{float}
\usepackage{amsfonts, amsmath, amsthm, amssymb}
\usepackage{caption}
\usepackage{tikz}
\usepackage{xcolor}
\usepackage{todonotes}
\usepackage[numbers]{natbib}
\usepackage{cleveref}
\Crefname{equation}{Eq.}{Eqs.}
\Crefname{figure}{Fig.}{Figs.}
\Crefname{tabular}{Tab.}{Tabs.}
\Crefname{section}{Sec.}{Secs.}
\Crefname{appendix}{App.}{Apps.}
\usepackage{wrapfig}

\newcommand*{\defeq}{\mathrel{\vcenter{\baselineskip0.5ex \lineskiplimit0pt
                     \hbox{\scriptsize.}\hbox{\scriptsize.}}}%
                     =}
\DeclareMathOperator*{\argmin}{argmin}

\usepackage[utf8]{inputenc}
\usepackage{intcalc}
\usepackage{ifthen}
\usetikzlibrary{patterns,arrows,calc,shapes}
\usetikzlibrary{decorations.markings,decorations.pathreplacing}

\newcommand{\myarrow}{-latex}

\title{The streaming rollout of deep networks - towards fully model-parallel execution}

\author{
  Volker Fischer \\
  Bosch Center for Artificial Intelligence \\
  Renningen, Germany \\
  \texttt{volker.fischer@de.bosch.com} \\
  \And
  Jan Köhler \\
  Bosch Center for Artificial Intelligence \\
  Renningen, Germany \\
  \texttt{jan.koehler@de.bosch.com} \\
  \And
  Thomas Pfeil \\
  Bosch Center for Artificial Intelligence \\
  Renningen, Germany \\
  \texttt{thomas.pfeil@de.bosch.com} \\
}

\begin{document}

\maketitle

\begin{abstract}
  Deep neural networks, and in particular recurrent networks, are promising candidates to control autonomous agents that interact in real-time with the physical world.
  However, this requires a seamless integration of temporal features into the network's architecture.
  For the training of and inference with recurrent neural networks, they are usually rolled out over time, and different rollouts exist.
  Conventionally during inference, the layers of a network are computed in a sequential manner resulting in sparse temporal integration of information and long response times.
  In this study, we present a theoretical framework to describe rollouts, the level of model-parallelization they induce, and demonstrate differences in solving specific tasks.
  We prove that certain rollouts, also for networks with only skip and no recurrent connections, enable earlier and more frequent responses, and show empirically that these early responses have better performance.
  The \emph{streaming} rollout maximizes these properties and enables a fully parallel execution of the network reducing runtime on massively parallel devices.
  Finally, we provide an open-source toolbox to design, train, evaluate, and interact with streaming rollouts.
\end{abstract}

\section{Introduction}
\label{sec:introduction}

\begin{wrapfigure}{R}{0.45\linewidth}
  \centering
  \input{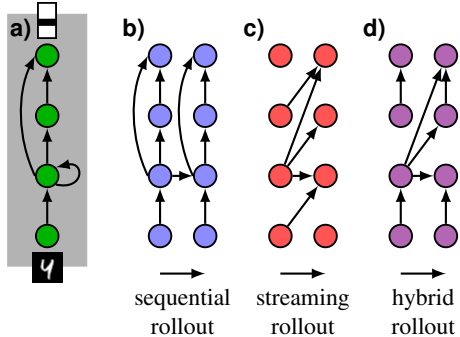}
  \caption{(best viewed in color) \textbf{a:} Neural network with skip and recurrent connections (SR) and different rollouts:
    \textbf{b:} the sequential rollout,
    \textbf{c:} the streaming rollout
    and \textbf{d:} a hybrid rollout.
    Nodes represent layers, edges represent transformations, e.g., convolutions.
    Only one rollout step is shown and each column in (b-d) is one frame within the rollout.
  }
  \label{fig:intro}
\end{wrapfigure}

Over the last years, the combination of newly available large datasets, parallel computing power, and new techniques to implement and train deep neural networks has led to significant improvements in the fields of vision \citep{He16}, speech \citep{amodei2016deep}, and reinforcement learning \citep{duan2016benchmarking}.
In the context of autonomous tasks, neural networks usually interact with the physical world in real-time
which renders it essential to integrate the processing of temporal information into the network's design.

Recurrent neural networks (RNNs) are one common approach to leverage temporal context and have gained increasing interest not only for speech \citep{fernandez2007application} but also for vision tasks \citep{graves2009offline}.
RNNs use neural activations to inform future computations, hence introducing a recursive dependency between neuron activations. 
This augments the network with a memory mechanism and allows it, unlike feed-forward neural networks, to exhibit dynamic behavior integrating a stream or sequence of inputs. 
For training and inference, backpropagation through time (BPTT) \citep{werbos1988generalization} or its truncated version \citep{werbos1988generalization, williams1995gradient} are used, where the RNN is \textit{rolled out} (or unrolled) through time disentangling the recursive dependencies and transforming the recurrent network into a feed-forward network.

Since unrolling a cyclic graph is not well-defined \citep{liao2016bridging}, different possible rollouts exist for the same neural network.
This is due to the rollout process itself, as there are several ways to unroll cycles with length greater \(1\) (larger cycles than recurrent self-connections).
More general, there are two ways to unroll every edge (cf. \Cref{fig:intro}): having the edge connect its source and target nodes at the same point in time (see, e.g., vertical edges in \Cref{fig:intro}b) or bridging time steps (see, e.g., \Cref{fig:intro}c).
Bridging is especially necessary for self-recurrent edges or larger cycles in the network, so that the rollout in fact becomes a feed-forward network. 
In a rollout, conventionally most edges are applied in the intra-frame non-bridging manner and bridge time steps only if necessary \citep{pascanu2013difficulty, liang2015recurrent, huang2015bidirectional, zamir2016feedback}. 
We refer to these rollouts as \textit{sequential} rollouts throughout this work.
One contribution of this study is the proof that the number of rollouts increases exponentially with network complexity.

The main focus of this work is that different rollouts induce different levels of model-parallelism and different behaviors for an unrolled network.
In rollouts inducing complete model-parallelism, which we call \textit{streaming}, nodes of a certain time step in the unrolled network become computationally disentangled and can be computed in parallel (see \Cref{fig:intro}c).
This idea is not restricted to recurrent networks, but generalizes to a large variety of network architectures covered by the presented graph-theoretical framework in \Cref{sec:theory}.
In \Cref{sec:experiments}, we show experimental results that emphasize the difference of rollouts for both, networks with recurrent and skip, and only skip connections.
In this study, we are not concerned comparing performances between networks, but between different rollouts of a given network 
(e.g., \Cref{fig:intro}b vs. c).

Our theoretical and empirical findings show that streaming rollouts enable fully model-parallel inference achieving low-latency and high-frequency responses.
These features are particularly important for real-time applications such as autonomous cars \citep{xu2017end} or UAV systems \citep{Lin2014Intelligent} in which the neural networks have to make complex decisions on high dimensional and frequent input signals within a short time.

To the best of our knowledge, up to this study, no general theory exists that compares different rollouts and our contributions can be summarized as follows:

\begin{itemize}
  \item We provide a theoretical framework to describe rollouts of deep neural networks and show that, and in some cases how, different rollouts lead to different levels of model-parallelism and network behavior.
  \item We formally introduce streaming rollouts enabling fully model-parallel network execution, and mathematically prove that streaming rollouts have the shortest response time to and highest sampling frequency of inputs.
  \item We empirically give examples underlining the theoretical statements and show that streaming rollouts can further outperform other rollouts by yielding better early and late performance.
  \item We provide an open-source toolbox specifically designed to study streaming rollouts of deep neural networks.
\end{itemize}

\section{Related work}\label{sec:existing_work}

The idea of RNNs dates back to the mid-70s \citep{little1974existence} and was popularized by \citep{hopfield1982neural}.
RNNs and their variants, especially Long Short-Term Memory networks (LSTM) \citep{Hochreiter97}, considerably improved performance in different domains such as speech recognition \citep{fernandez2007application}, handwriting recognition \citep{graves2009offline}, machine translation \citep{sutskever2014sequence}, optical character recognition (OCR) \citep{breuel2013high}, text-to-speech synthesis \citep{fan2014tts}, social signal classification \citep{brueckner2014social}, or online multi-target tracking \citep{milan2017online}. 
The review \citep{schmidhuber2015deep} gives an overview of the history and benchmark records set by DNNs and RNNs.

\textbf{Variants of RNNs:}
There are several variants of RNN architectures using different mechanisms to memorize and integrate temporal information. These include LSTM networks \citep{Hochreiter97} and related architectures like Gated Recurrent Unit (GRU) networks \citep{cho2014properties} or recurrent highway networks \citep{zilly2016recurrent}. 
Neural Turing Machines (NTM) \citep{graves2014neural} and Differentiable Neural Computers (DNC) \citep{graves2016hybrid} 
extend RNNs by an addressable external memory. 
Bi-directional RNNs (BRNNs) \citep{schuster1997bidirectional} incorporate the ability to model the dependency on future information.
Numerous works extend and improve these RNN variants creating architectures with advantages for training or certain data domains 
(e.g., \citep{campos2018skip, pundak2017highway, pascanu2013construct, chung2017hierarchical}).

\textbf{Response time:}
While RNNs are the main reason to use network rollouts, in this work we also investigate rollouts for non-recurrent networks.
Theoretical and experimental results suggest that different rollout types yield different behavior especially for networks containing skip connections.
The rollout pattern influences the \emph{response time} of a network which is the duration between input (stimulus) onset and network output (response).

Shortcut or skip connections can play an important role to decrease response times. 
Shortcut branches attached to intermediate layers allow earlier predictions (e.g., BranchyNet \citep{Teerapittayanon16}) and iterative predictions refine from early and coarse to late and fine class predictions (e.g., feedback networks \citep{zamir2016feedback}). 
In \citep{Greff17}, the authors show that identity skip connections, as used in Residual Networks (ResNet) \citep{He16}, can be interpreted as local network rollouts acting as filters, which could also be achieved through recurrent self-connections. 
The good performance of ResNets underlines the importance of local recurrent filters.
The runtime of inference and training for the same network can also be reduced by network compression \citep{han2015deep, kim2015compression} or optimization of computational implementations \citep{lavin2016fast, mathieu2014fast}.

\textbf{Rollouts:}
To train RNNs, different rollouts are applied in the literature, though lacking a theoretically founded background.
One of the first to describe the transformation of a recurrent MLP into an equivalent feed-forward network and depicting it in a streaming rollout fashion was \citep[ch. 9.4]{minsky1969perceptrons}. 
The most common way in literature to unroll networks over time is to duplicate the model for each time step as depicted in \Cref{fig:intro}b \citep[ch. 10.1 in][]{goodfellow2016deep, pascanu2013difficulty, liang2015recurrent, huang2015bidirectional, zamir2016feedback, Gregor2018}.
However, as we will show in this work, this rollout pattern is neither the only way to unroll a network nor the most efficient.

The recent work of \citet{Carreira2018} also addresses the idea of model-parallelization through dedicated network rollouts 
to reduce latency between input and network output by distributing computations over multiple GPUs.
While their work shows promising empirical findings in the field of video processing, 
our work provides a theoretical formulation for a more general class of networks and their rollouts.
Our work also differs in the way the delay between input and output, 
and network training is addressed.

Besides the chosen rollout, other methods exist, that modify the integration of temporal information:
for example, \textit{temporal stacking} (convolution over time), which imposes a fixed temporal receptive field 
(e.g., \citep{Tran2015Spatiotemporal, Heilbron2015ActivityNet}), 
\textit{clocks}, where different parts of the network have different update frequencies, 
(e.g., \citep{Koutnik2014Clockwork, Vezhnevets2017Feudal, Shelhamer2016Clockwork, Figurnov2017Spatially, Neil16})
or \textit{predictive states}, which try to compensate temporal delays between different network parts (e.g., \citep{Carreira2018}). For more details, please see also \Cref{sec:discussion}.

\section{Graph representations of network rollouts}
\label{sec:theory}

We describe dependencies inside a neural network \(N\) as a directed graph \(N = (V,E)\). 
The nodes \(v \in V\) represent different layers and the edges \(e \in E \subset V \times V\) represent transformations introducing direct dependencies between layers. 
We allow self-connections \((v,v) \in E\) and larger cycles in a network. 
Before stating the central definitions and propositions, 
we introduce notations used throughout this section and for the proofs in the appendix.

Let \(G=(V,E)\) be a directed graph with vertices (or nodes) \(v \in V\) and edges \(e = (e_\text{src},e_\text{tgt}) \in E \subset V \times V\). 
Since neural networks process input data, we denote the \textbf{input} of the graph as set \(I_G\), consisting of all nodes without incoming edges:
\begin{equation}\label{eq:definitionpath}
	I_G \defeq \{ v \in V \ | \ \nexists u \in V : (u,v) \in E\}.
\end{equation}
A \textbf{path} in \(G\) is a mapping \(p:\{1, \ldots, L\} \rightarrow E\) with \(p(i)_\text{tgt} = p(i+1)_\text{src}\) for \(i \in \{1, \ldots, L - 1\}\) where \(L \in \mathbb{N}\) is the \textbf{length} of \(p\). 
We denote the length of a path \(p\) also as \(|p|\) and the number of elements in a set \(A\) as \(|A|\). 
A path \(p\) is called \textbf{loop} or \textbf{cycle} iff \(p(|p|)_\text{tgt} = p(1)_\text{src}\) and it is called \textbf{minimal} iff \(p\) is injective.
The set of all cycles is denoted as \(C_G\).
Two paths are called \textbf{non-overlapping} iff they share no edges.
We say a graph is \textbf{input-connected} iff for every node \(v\) exists a path \(p\) with \(p(|p|)_\text{tgt}=v\) and \(p(1)_\text{src} \in I_G\).
Now we proceed with our definition of a (neural) network.

\paragraph{Definition (network):}\label{def:network} A \textbf{network} is a directed and input-connected graph \(N=(V,E)\) for which \(0 < |E| < \infty\).

\begin{figure}
	\centering
  \input{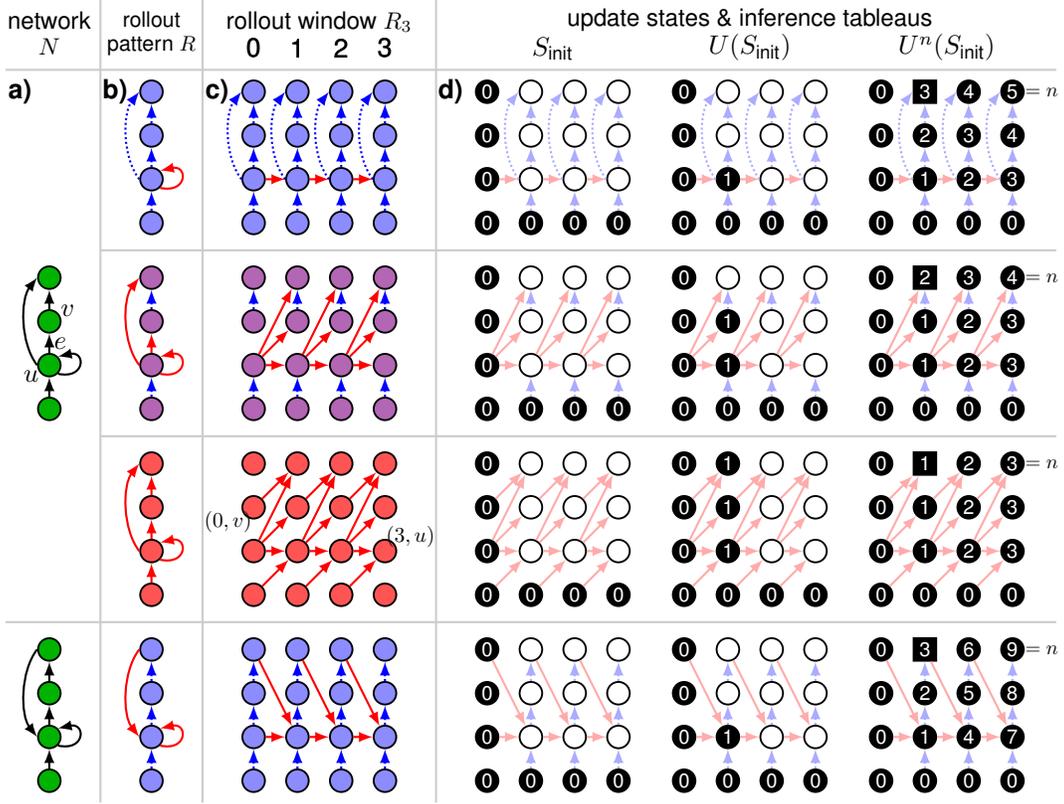}
  \caption{
    (best viewed in color) \textbf{a:} Two different networks. 
    \textbf{b:} Different rollout patterns \(R:E \rightarrow \{0,1\}\) for the two networks.
    \textit{Sequential} (\(R(e) = 0\)) and \textit{streaming} (\(R(e) = 1\)) edges are indicated with blue dotted and red solid arrows respectively. 
    For the first network (top to bottom), the most sequential, one hybrid, and the streaming rollout patterns are shown. 
    For the second network, one out of its \(3\) most sequential rollout patterns is shown (either of the three edges of the cycle could be unrolled).
    \textbf{c:} Rollout windows of size \(W = 3\). 
    By definition (\Cref{eq:def_rollout}), sequential and streaming edges propagate information within and to the next frames respectively. 
    \textbf{d:} States \(S(\overline{v})\) and inference tableau values \(T(\overline{v})\). 
    The state \(S(\overline{v})\) of a node is indicated with black (already known) or white (not yet computed). 
    From left to right: initial state \(S_{\text{init}}\),
    state after first update step \(U(S_{\text{init}})\), full state \(U^n(S_{\text{init}}) = S_{\text{full}}\). 
    The number of update steps \(n\) to reach the full state differs between rollouts. 
    Numbers inside nodes \(\overline{v}\) indicate values of the inference tableau (\(T(\overline{v})\)).
    Inference factors \(F(R)\) are indicated with square instead of circular nodes in the first frame of the full states.
  }
  \label{fig:theory}
\end{figure}

For our claims, this abstract formulation is sufficient and, while excluding certain artificial cases, it ensures that a huge variety of neural network types is covered (see \Cref{fig:app_network_examples} for network examples).
For deep neural networks, we give an explicit formulation of this abstraction in \Cref{sec:theory_dnn}, which we also use for our experiments. Important concepts introduced here are illustrated in \Cref{fig:theory}.
In this work, we separate the concept of network rollouts into two parts: The temporal propagation scheme which we call \emph{rollout pattern} and its associated \emph{rollout windows} (see also \Cref{fig:intro} and \Cref{fig:theory}):

\paragraph{Definition (rollout pattern and window):}\label{def:rollout}
Let \(N=(V,E)\) be a network.
We call a mapping \(R: E \rightarrow \{0, 1\}\) a \textbf{rollout pattern} of \(N\).
For a rollout pattern \(R\), the \textbf{rollout window} of size \(W\in\mathbb{N}\) is the directed graph \(R_W=(V_W, E_W)\) with:
\begin{equation}\label{eq:def_rollout}
\begin{split}
	V_W & \defeq \{0, \ldots, W\} \times V,\ \ \ \overline{v} = (i,v) \in V_W \\
	E_W & \defeq \{((i,u),(j,v)) \in V_W \times V_W \ \ | \ \ (u,v) \in E \ \land \ j = i + R((u,v))\}.
\end{split}
\end{equation}
Edges \(e \in E\) with \(R(e) = 1\) enable information to directly \emph{stream} through time.
In contrast, edges with \(R(e) = 0\) cause information to be processed within frames, thus introducing \emph{sequential} dependencies upon nodes inside a frame.
We dropped the dependency of \(E_W\) on the rollout pattern \(R\) in the notation.
A rollout pattern and its rollout windows are called \textbf{valid} iff \(R_W\) is acyclic for one and hence for all \(W \in \mathbb{N}\).
We denote the set of all valid rollout patterns as \(\mathcal{R}_{N}\) and the rollout pattern \(R \equiv 1\) the \textbf{streaming rollout} \(R^\text{stream} \in \mathcal{R}_N\). We say two rollout patterns \(R\) and \(R^{\prime}\) are \textbf{equally model-parallel} 
iff they are equal (\(R(e) = R^{\prime}(e)\)) for all edges \(e = (u,v) \in E\), not originating in the network's input (\(u \notin I_N\)).
For \(i \in \{0, \ldots, W\}\), the subset \(\{i\} \times V \subset V_W\) is called the \(i\)-th \textbf{frame}.

\textbf{Proof:} In \Cref{sec:proof_rollout}, we prove that the definition of valid rollout patterns is well-defined and is consistent with intuitions about rollouts, such as consistency over time.
We also prove that the streaming rollout exists for every network and is always valid.

The most non-streaming rollout pattern \(R \equiv 0\) is not necessarily valid, because if \(N\) contains loops then \(R \equiv 0\) does not yield acyclic rollout windows.
Commonly, recurrent networks are unrolled such that most edges operate inside the same frame (\(R(e) = 0\)), and only when necessary (e.g., for recurrent or top-down) connections are unrolled (\(R(e) = 1\)).
In contrast to this sequential rollout, the streaming rollout pattern unrolls all edges with \(R(e)=1\) (cf. top and third row in \Cref{fig:theory}).

\paragraph{Lemma 1:}\label{lemma1} Let \(N=(V,E)\) be a network. 
The number of valid rollout patterns \(|\mathcal{R}_{N}|\) is bounded by:
\begin{equation}\label{eq:lemma1}
	1 \leq n \leq |\mathcal{R}_{N}| \leq 2^{|E| - |E_{\text{rec}}|},
\end{equation}
where \(E_{\text{rec}}\) is the set of all self-connecting edges \(E_{\text{rec}} \defeq \{(u, v) \in E\ |\ u = v\}\), and \(n\) either:
\begin{itemize}
	{\item \(n = 2^{|E_\text{forward}|}\), with \(E_\text{forward}\) being the set of edges not contained in any cycle of \(N\), or
	\item \(n = \prod\limits_{p \in C}(2^{|p|} - 1)\), \(C \subset C_N\) being any set of minimal and pair-wise non-overlapping cycles.}
	\vspace*{-\baselineskip}
\end{itemize}
\textbf{Proof:} See appendix \Cref{sec:proof_lemma1}.

Lemma 1 shows that the number of valid rollout patterns increases exponentially with network complexity. 
Inference of a rollout window is conducted in a sequential manner.
This means, the state of all nodes in the rollout window is successively computed depending on the availability of already computed source nodes\footnote{given the state of all input nodes at all frames and initial states for all nodes at the zero-th frame}.
The chosen rollout pattern determines the mathematical function this rollout represents, which may be different between rollouts, e.g., for skip connections.
In addition, the chosen rollout pattern also determines the order in which nodes can be computed leading to different runtimes to compute the full state of a rollout window.

We now introduce tools to compare these addressed differences between rollouts.
\textit{States} of the rollout window encode, which nodes have been computed so far and \textit{update steps} determine the next state based on the previous state.
\textit{Update tableaus} list after how many update steps nodes in the rollout window are computed. 
Update states, update steps, and inference tableaus are shown for example networks and rollouts in \Cref{fig:theory}.

\paragraph{Definition (update state, update step, tableau, and factor):} \label{def:state_step_tableau}
Let \(R\) be a valid rollout pattern of a network \(N=(V,E)\).
A \textbf{state} of the rollout window \(R_W\) is any mapping \(S:V_W \rightarrow \{0, 1\}\).
Let $\Sigma_W$ denote the set of all possible states.
We define the \textbf{full state} \(S_\text{full}\) and \textbf{initial state} \(S_\text{init}\) as:
\begin{equation}\label{eq:examplesS}
	S_\text{full} \equiv 1; \ \ \ \ \ \ \ \ S_\text{init}((i,v)) = 1\ \Longleftrightarrow\ v \in I_{N} \lor i=0.
\end{equation}
Further, we define the \textbf{update step} \(U\) which updates states \(S\).
Because the updated state \(U(S)\) is again a state and hence a mapping, we define \(U\) by specifying the mapping \(U(S)\):
\begin{equation}\label{eq:definitionU}
	U:\Sigma_W \rightarrow \Sigma_W;\ \ \ \ \ \ \ \ U(S): V_W \rightarrow \{0,1\}
\end{equation}
\[
U(S)(\overline{v}) \defeq
\left \{
  \begin{tabular}{cl}
  \(1\) & if \(S(\overline{v})=1\) or if for all \((\overline{u},\overline{v}) \in E_W \ : \ S(\overline{u}) = 1\) \\
  \(0\) & otherwise
  \end{tabular}
\right.
\]
We call the mapping \(T : V_W \rightarrow \mathbb{N}\) the \textbf{inference tableau}:
\begin{equation}\label{eq:tableau}
	T(\overline{v}) \ \ \defeq \ \ \max\limits_{p \in P_{\overline{v}}} |p| \ \ = \ \ \argmin_{n\in\mathbb{N}}\left\{U^{n}(S_\text{init})(\overline{v}) = 1 \right\} 
\end{equation}
where \(U^n\) is the \(n\)-th recursive application of \(U\) and for \(\overline{v} \in V_W\), \(P_{\overline{v}}\) denotes the set of all paths in \(R_W\) that end at \(\overline{v}\) (i.e., \(p(|p|)_\text{tgt} = \overline{v}\)) and for which their first edge may start but not end in the \(0\)-th frame, \(p(1)_\text{tgt} \notin \{0\}\times V\). Hereby, we exclude edges (computational dependencies) which never have to be computed, because all nodes in the \(0\)-th frame are initialized from start. 
We dropped the dependencies of \(U\) and \(T\) on the rollout window \(R_W\) in the notation and if needed we will express them with \(U_{R_W}\) and \(T_{R_W}\).
Further, we call the maximal value of \(T\) over the rollout window of size \(1\) the rollout pattern's \textbf{inference factor}:
\begin{equation}\label{eq:factor}
	F(R) \defeq \max\limits_{\overline{v} \in V_1} T_{R_1}(\overline{v}).
\end{equation}
\textbf{Proof:} In \Cref{sec:proof_inference_update_tableau} we prove \Cref{eq:tableau}.

\label{theory:u_conv}
We also want to note that all rollout windows of a certain window size \(W\) have the same number of edges \(W * |E|\), independent of the chosen rollout pattern (ignoring edges inside the \(0\)-th frame, because these are not used for updates). However, maximal path lengths in the rollout windows differ between different rollout patterns (cf. \Cref{eq:tableau} and its proof, as well as tableau values in \Cref{fig:theory}).

Inference of rollout windows starts with the initial state \(S_\text{init}\).
Successive applications of the update step \(U\) updates all nodes until the fully updated state \(S_\text{full}\) is reached (cf. \Cref{fig:theory} and see \Cref{sec:proof_u_conv} for a proof).
For a certain window size \(W\), the number of operations to compute the full state is independent of the rollout pattern, but which updates can be done in parallel heavily depends on the chosen rollout pattern.
We will use the number of required update steps to measure computation time. 
This number differs between different rollout patterns (e.g., \(F(R)\) in \Cref{fig:theory}).
In practice, the time needed for the update \(U(S)\) of a certain state \(S\) depends on \(S\) 
(i.e., which nodes can be updated next).
For now, we will assume independence, but will address this issue in the discussion (\Cref{sec:discussion}).

\paragraph{Theorem 1:} \label{theorem:pip}
Let \(R\) be a valid rollout pattern for a network \(N=(V,E)\) then the following statements are equivalent:
\vspace*{-0.5\baselineskip}
\begin{itemize}
{
	\item[a)] \(R\) and the streaming rollout pattern \(R^\text{stream}\) are equally model-parallel.
	\item[b)] The first frame is updated entirely after the first update step: \(F(R) = 1\).
  	\item[c)] For \(W\in\mathbb{N}\), the \(i\)-th frame of \(R_W\) is updated at the \(i\)-th update step:
\[
  \forall (i,v) \in V_W \ : \  T((i,v)) \leq i.
\]
	\item[d)] For \(W\in\mathbb{N}\), the inference tableau of \(R_W\) is minimal everywhere and over all rollout patterns.
    In other words, responses are earliest and most frequent:
\[
	\forall \overline{v} \in V_W \ :\ T_{R_W}(\overline{v}) = \min\limits_{R^{\prime} \in \mathcal{R}_N}T_{R^{\prime}_W}(\overline{v}).
\]
}
\vspace*{-1em}
\end{itemize}
\vspace*{-0.5\baselineskip}
\textbf{Proof:} See appendix \Cref{sec:proof_pip}.

\section{Experiments}
\label{sec:experiments}

\begin{figure}
  \centering
  \includegraphics[width=0.95\linewidth]{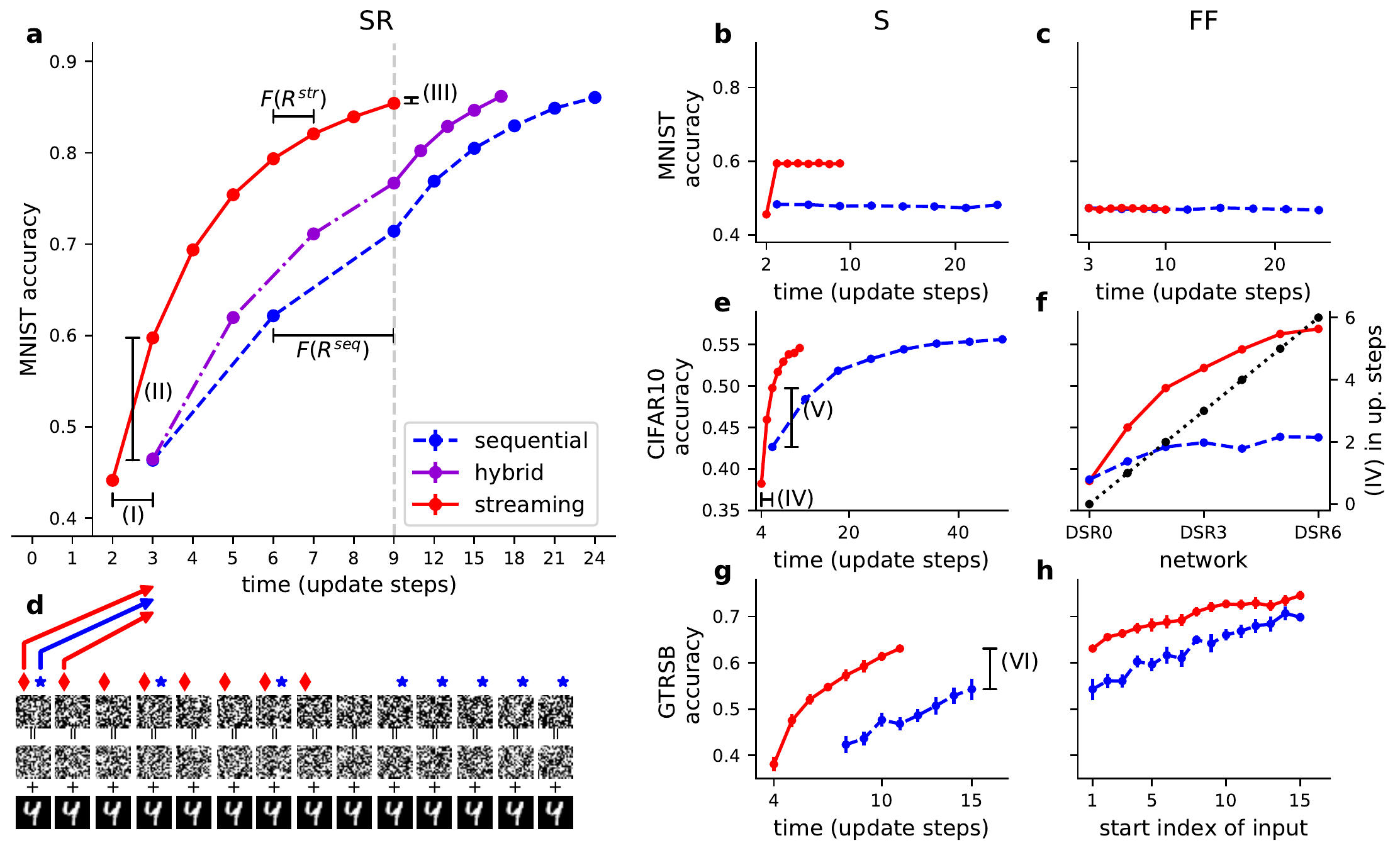}
  \caption{
    (best viewed in color) Classification accuracy for sequential (in dashed blue), streaming (in solid red), and one hybrid (violet; only for SR network in a) rollout on MNIST, CIFAR10, and GTSRB (for networks and data see Figs. \ref{fig:intro}, \ref{fig:results}d, \ref{fig:methods}, and \ref{fig:app_network_mnist}).
    \textbf{a-c:} Average classification results on MNIST over computation time measured in the number of update steps of networks with skip + recurrent (SR, a), with skip (S, b), and only feed-forward (FF, c) connections. 
    In a), scaling of the abscissa changes at the vertical dashed line for illustration purposes.
    \textbf{d:} The input (top row) is composed of digits (bottom row) and noise (middle row).
    Note that the input is aligned to the time axis in (a).
    Red diamonds and blue stars indicate inputs sampled by streaming and sequential rollouts, respectively.
    \textbf{e:} Classification results of the network DSR2 on CIFAR10.
    \textbf{f:} Accuracies at time of first output of sequential rollout (see (V) in e) over networks DSR0 - DSR6 (red and blue curves; left axis). Differences of first response times between streaming and sequential rollouts (see (IV) in e; black dotted curve; right axis).
    \textbf{g:} Average accuracies on GTSRB sequences starting at index $0$ of the original sequences.
    \textbf{h:} Final classification accuracies (see (VI) in g) over the start index of the input sequence.
    Standard errors are shown in all plots except e and f and are too small to be visible in (a-c).
  }
  \label{fig:results}
\end{figure}

To demonstrate the significance of the chosen rollouts w.r.t. the runtime for inference and achieved accuracy, 
we compare the two extreme rollouts: the most model-parallel, i.e., streaming rollout (\(R \equiv 1\), results in red in \Cref{fig:results}), and the most sequential rollout\footnote{Here, the most sequential rollout is unique since the used networks do not contain cycles of length greater \(1\). For sequential rollouts that are ambiguous see bottom row of \Cref{fig:theory}.} (\(R(e) = 0\) for maximal number of edges, results in blue in \Cref{fig:results}).

In all experiments, we consider a response time task, in which the input is a sequence of images and the networks have to respond as quickly as possible with the correct class. 
We want to restate that we do not compare performances between networks but between rollout patterns of the same network.

For all experiments and rollout patterns under consideration, we conduct inference on shallow rollouts (\(W = 1\)) 
and initialize the zero-th frame of the next rollout window with the last (i.e., \(1.\)) frame of the preceding rollout window (see discussion \Cref{sec:discussion}). 
Hence, the inference factor of a rollout pattern is used to determine the number of update steps between responses (see \(F(R^{\text{str}})\), \(F(R^{\text{seq}})\) in \Cref{fig:results}a).

\textbf{Datasets:}
Rollout patterns are evaluated on three datasets: MNIST \citep{lecun1998mnist}, CIFAR10 \citep{krizhevsky2009learning}, and the German traffic sign recognition benchmark (GTSRB) \citep{stallkamp2011german}.
To highlight the differences between different rollout patterns, we apply noise (different sample for each frame) to the data 
(see \Cref{fig:results}d and \Cref{fig:methods}b, c).
In contrast to data without noise, a single image is now not sufficient for a good classification performance anymore and 
temporal integration is necessary.
In case of GTSRB, this noise can be seen as noise induced by the sensor as predominant under poor lighting conditions.
GTSRB contains tracks of \(30\) frames from which sections are used as input sequences.

\textbf{Networks:}
We compare the behavior of streaming and sequential rollout patterns on MNIST for three different networks with two hidden layers (FF, S, SR; see \Cref{fig:intro} and \Cref{fig:app_network_mnist}).
For evaluation on CIFAR10, we generate a sequence of \(7\) incrementally deeper networks (DSR0 - DSR6, see \Cref{fig:methods}a) by adding layers to the blocks of a recurrent network with skip connections in a dense fashion (details in \Cref{fig:methods}a).
For evaluation on GTSRB, we used DSR4 leaving out the recurrent connection.
Details about data, preprocessing, network architectures, and the training process are given in \Cref{sec:details_nets}.

\textbf{Results:}
Rollouts are compared on the basis of their test accuracies over the duration (measured in update steps) needed to achieve these accuracies (\Cref{fig:results}a-c, e, and g). 

We show behavioral differences between streaming and sequential rollouts for increasingly complex networks on the MNIST dataset.
In the case of neither recurrent, nor skip connections (see FF in \Cref{fig:app_network_mnist}), the streaming rollout is mathematically identical to the sequential rollout.
Neither rollout can integrate information over time and, hence, both perform classification on single images with the same response time for the first input image and same accuracy (see \Cref{fig:results}c).
However, due to the pipelined structure of computations in the streaming case, outputs are more frequent.

For networks with skip, but without recurrent connections (see S in \Cref{fig:app_network_mnist}), 
the behavioral difference between streaming and sequential rollouts can be shown best.
While the sequential rollout still only performs classification on single images, the streaming rollout can integrate over two input images due to the skip connection that bridges time (see \Cref{fig:results}b).

In the streaming case, skip connections cause shallow shortcuts in time that can result in earlier (see (I) in \Cref{fig:results}a), but initially worse performance than for deep sequential rollouts.
The streaming rollout responds $1$ update step earlier than the sequential rollout since its shortest path is shorter by $1$ (see \Cref{fig:theory}).
These early first estimations are later refined when longer paths and finally the longest path from input to output contribute to classification.
For example, after $3$ time steps in \Cref{fig:results}a, the streaming rollout uses the full network.
This also applies to the sequential rollout, but instead of integrating over two images (frames \(0\) and \(1\)), only the image of a single frame (frame \(1\)) is used (cf. blue to red arrows connecting \Cref{fig:results}d and a).

Due to parallel computation of the entire frame in the streaming case, the sampling frequency of input images (every time step; see red diamonds in \Cref{fig:results}d) is maximal (\(F(R^{\text{str}}) = 1\) in \Cref{fig:results}a; see d in Theorem 1 in \Cref{theorem:pip}).
In contrast, the sampling frequency of the sequential rollout decreases linearly with the length of the longest path (\(F(R^{\text{seq}}) = 3\) in \Cref{fig:results}a; blue stars in \Cref{fig:results}d).

High sampling frequencies and shallow shortcuts via skip connections establish a high degree of temporal integration early on and result in better early performance (see (II) in \Cref{fig:results}a).
In the long run, however, classification performances are comparable between streaming and sequential rollouts and the same number of input images is integrated over (see (III) in \Cref{fig:results}a).

We repeat similar experiments for the CIFAR10 dataset to demonstrate the increasing advantages of the streaming over sequential rollouts for deeper and more complex networks.
For the network DSR2 with the shortest path of length \(4\) and longest path of length \(6\), the first response of the streaming rollout is $2$ update steps earlier than for the sequential rollout (see (IV) in \Cref{fig:results}e) and shows better early performance (see (V) in \Cref{fig:results}e).
With increasing depth (length of the longest path) over the sequence of networks DSR0 - DSR6 (see  \Cref{fig:methods}a), the time to first response stays constant for streaming, but linearly grows with the depth for sequential rollouts (see \Cref{fig:results}f black curve).
The difference of early performance (see (V) in \Cref{fig:results}e) widens with deeper networks (\Cref{fig:results}f).

For evaluation of rollouts on GTSRB, we considere the DSR4 network.
Self-recurrence is omitted since the required short response times of this task cannot be achieved with sequential rollouts due to the very small sampling frequencies.
Consequently, for fair comparison, we calculate the classifications of the first $8$ images in parallel for the sequential case.
In this case, where both rollouts use the same amount of computations, performance for the sequential rollout increases over time due to less blurry input images, while the streaming rollout in addition performs temporal integration using skip connections and yields better performance (see (VI) in \Cref{fig:results}g).
This results in better performance of streaming compared to sequential rollouts for more distant objects (\Cref{fig:results}h).

\section{Discussion and Conclusion}
\label{sec:discussion}

The presented theory for network rollouts is generically applicable to a vast variety of deep neural networks (see \Cref{sec:theory_dnn}) and is not constrained to recurrent networks but could be used on forward (e.g., VGG \citep{simonyan2014very}, AlexNet \citep{krizhevsky2012imagenet}) or skipping networks (e.g., ResNet \citep{He16}, DenseNet \citep{huang2017densely}).
We restricted rollout patterns to have values \(R(e) \in \{0,1\}\) and did neither allow edges to bridge more than \(1\) frame \(R(e) > 1\) nor pointing backwards in time \(R(e) < 0\). 
The first case is subsumed under the presented theory using copy-nodes for longer forward connections,
and for \(R(e) < 0\) rollouts with backward connections loose the real-time capability, because information from future frames would be used.

In this work, we primarily investigated differences between rollout patterns in terms of the level of parallelization they induce in their rollout windows.
But using different rollout patterns is not a mere implementation issue. 
For some networks, all rollout patterns yield the same mathematical behavior (e.g., mere feed-forward networks without any skip or recurrent connections, cf. \Cref{fig:results}c). 
For other networks, different rollout patterns (see \Cref{def:network}) may lead to differences in the behavior of their rollout windows (e.g., \Cref{fig:results}b). 
Hence, parameters between different rollout patterns might be incompatible. The theoretical analysis of \textit{behavioral} equivalency of rollout patterns is a topic for future work.

One disadvantage of the streaming rollout pattern seems to be that deeper networks also require deeper rollout windows.
Rollout windows should be at least as long as the longest minimal path connecting input to output, 
i.e., all paths have appeared at least once in the rollout window. 
For sequential rollout patterns this is not the case, since, 
e.g., for a feed-forward network the longest minimal path is already contained in the first frame.
However, for inference with streaming rollouts instead of using deep rollouts we propose to use shallow rollouts (e.g., \(W = 1\))
and to initialize the zero-th frame of the next rollout window with the last (i.e., first) frame of the preceding rollout window.
This enables a potentially infinite memory for recurrent networks and minimizes the memory footprint of the rollout window during inference.

Throughout the experimental section, we measured runtime by the number of necessary update steps assuming equal update time for every single node update.
Without this assumption and given fully parallel hardware, 
streaming rollouts still manifest the best case scenario in terms of maximal parallelization and the inference of a single frame would take the runtime of the computationally most expensive node update. 
However, sequential rollouts would not benefit from the assumed parallelism of such hardware and inference of a single frame takes the summed up runtime of all necessary node updates.
The streaming rollout favors network architectures with many nodes of approximately equal update times. 
In this case, the above assumtion approximately holds.

The difference in runtime between rollout patterns depends on the hardware used for execution.
Although commonly used GPUs provide sufficient parallelism to speed up calculations of activations within a layer, they are often not parallel enough to enable the parallel computation of multiple layers.
Novel massively parallel hardware architectures such as the TrueNorth chip \citep{Merolla_2014_668, Esser_2016_11441} allow to store and run the full network rollouts on-chip reducing runtime of rollouts drastically and therefore making streaming rollouts highly attractive.
The limited access to massively parallel hardware may be one reason, why streaming rollouts have not been thoroughly discussed, yet.

Furthermore, not only the hardware, but also the software frameworks must support the parallelization of independent nodes in their computation graph to exploit the advantages of streaming rollouts.
This is usually not the case and by default sequential rollouts are used.
For the experiments presented here, we use the Keras toolbox to compare different rollout patterns. 
To realize arbitrary rollout patterns in Keras, instead of using Keras' build-in RNN functionalities, 
we created a dedicated model builder which explicitly generates the rollout windows.
Additionally, we implemented an experimental toolbox (Tensorflow and Theano backends) to study (define, train, evaluate, and visualize) networks using the streaming rollout pattern (see \Cref{sec:statestream_toolbox}).
Both are available as open-source code\footnote{\url{https://github.com/boschresearch/statestream}}.

Similar to biological brains, synchronization of layers (nodes) plays an important role for the streaming rollout pattern.
At a particular time (frame), different nodes may carry differently delayed information with respect to the input. 
In this work, we evaluate network accuracy dependent on the delayed response.
An interesting area for future research is the exploration of mechanisms to guide and control information flow in the context of the streaming rollout patterns, e.g., through gated skips (bottom-up) and recurrent (top-down) connections.
New sensory information should be distributed quickly into deeper layers. 
and high-level representations and knowledge of the network about its current task could stabilize, 
predict, and constrain lower-level representations.

A related concept to layer synchronization is that of \textit{clocks}, where different layers, or more generally different parts of a network,
are updated with different frequencies. 
In this work, all layers are updated equally often. In general, it is an open research question to which extend clocking and more generally synchronization mechanisms should be implicit parts of the network and hence learnable or formulated as explicit a-priory constraints.

\textbf{Conclusion:}
We presented a theoretical framework for network rollouts and investigated differences in behavior and model-parallelism between different rollouts. We especially analysed the streaming rollout, which fully disentangles computational dependencies between nodes and hence enables full model-parallel inference. We empirically demonstrated the superiority of the streaming over non-streaming rollouts for different image datasets due to faster first responses to and higher sampling of inputs.
We hope our work will encourage the scientific community to further study the advantages and behavioral differences of streaming rollouts in preparation to future massively parallel hardware.

\clearpage
\subsubsection*{Acknowledgments}

The authors would like to thank Bastian Bischoff, Dan Zhang, Jan-Hendrik Metzen, and Jörg Wagner for their valuable remarks and discussions.

{\small
\bibliographystyle{unsrtnat} 
\bibliography{bibtex_references}
}

\clearpage
\section*{The streaming rollout of deep networks - towards fully model-parallel execution \\ Supplementary material}

\renewcommand{\thetable}{A\arabic{table}}
\renewcommand{\thefigure}{A\arabic{figure}}
\renewcommand{\thesection}{A\arabic{section}}
\setcounter{table}{0}
\setcounter{figure}{0}
\setcounter{section}{0}
\setcounter{page}{1}

\section{Proofs and notes for theory chapter}
\label{sec:proofs}

To improve readability, we will restate certain parts of the theory chapter from the main text.

\subsection{Examples for networks following our definition in \Cref{def:network}}

\begin{figure}[H]
	\centering
  \def\NodeSize{3.0}
\def\NodeBorder{0.25}

\begin{tikzpicture}

	\draw (1.5, 0.5) node[draw, circle, minimum size=\NodeSize mm, inner sep=0pt, fill=black!30!green, line width=\NodeBorder mm] (vert11) {};
	\draw (1.5, 1.3) node[draw, circle, minimum size=\NodeSize mm, inner sep=0pt, fill=black!30!green, line width=\NodeBorder mm] (vert12) {};
	\draw (1.5, 2.1) node[draw, circle, minimum size=\NodeSize mm, inner sep=0pt, fill=black!30!green, line width=\NodeBorder mm] (vert13) {};
	\draw[\myarrow, thick] (vert11) -- (vert12);
	\draw[\myarrow, thick] (vert12) -- (vert13);

	\draw (2.5, 0.5) node[draw, circle, minimum size=\NodeSize mm, inner sep=0pt, fill=black!30!green, line width=\NodeBorder mm] (vert21) {};
	\draw (2.5, 1.3) node[draw, circle, minimum size=\NodeSize mm, inner sep=0pt, fill=black!30!green, line width=\NodeBorder mm] (vert22) {};
	\draw (2.5, 2.1) node[draw, circle, minimum size=\NodeSize mm, inner sep=0pt, fill=black!30!green, line width=\NodeBorder mm] (vert23) {};
	\draw[\myarrow, thick] (vert21) -- (vert22);
	\draw[\myarrow, thick] (vert22.north west) -- (vert23.south west);
	\draw[\myarrow, thick] (vert23.south east) -- (vert22.north east);
	\draw[\myarrow, thick] (vert21.west) .. controls (2.0, 1.0) and (2.0, 1.6) .. (vert23.west);
	\draw[\myarrow, thick] (vert22.south east) .. controls (3.0, 1.1) and (3.0, 1.5) .. (vert22.north east);

	\draw (3.5, 0.9) node[draw, circle, minimum size=\NodeSize mm, inner sep=0pt, fill=black!30!green, line width=\NodeBorder mm] (vert31) {};
	\draw (3.5, 1.8) node[draw, circle, minimum size=\NodeSize mm, inner sep=0pt, fill=black!30!green, line width=\NodeBorder mm] (vert32) {};
	\draw (4.4, 0.9) node[draw, circle, minimum size=\NodeSize mm, inner sep=0pt, fill=black!30!green, line width=\NodeBorder mm] (vert33) {};
	\draw (4.4, 1.8) node[draw, circle, minimum size=\NodeSize mm, inner sep=0pt, fill=black!30!green, line width=\NodeBorder mm] (vert34) {};
	\draw[\myarrow, thick] (vert31) -- (vert32);
	\draw[\myarrow, thick] (vert32) -- (vert34);
	\draw[\myarrow, thick] (vert34) -- (vert33);
	\draw[\myarrow, thick] (vert33) -- (vert31);
	\draw[red, line width=1mm] (4.2, 0.5) -- (3.7, 2.2);

	\draw (5.5, 1.3) node[draw, circle, minimum size=\NodeSize mm, inner sep=0pt, fill=black!30!green, line width=\NodeBorder mm] (vert41) {};
	\draw (5.5, 2.1) node[draw, circle, minimum size=\NodeSize mm, inner sep=0pt, fill=black!30!green, line width=\NodeBorder mm] (vert42) {};
	\draw (6.4, 1.3) node[draw, circle, minimum size=\NodeSize mm, inner sep=0pt, fill=black!30!green, line width=\NodeBorder mm] (vert43) {};
	\draw (6.4, 2.1) node[draw, circle, minimum size=\NodeSize mm, inner sep=0pt, fill=black!30!green, line width=\NodeBorder mm] (vert44) {};
	\draw (5.95, 0.5) node[draw, circle, minimum size=\NodeSize mm, inner sep=0pt, fill=black!30!green, line width=\NodeBorder mm] (vert45) {};
	\draw[\myarrow, thick] (vert41) -- (vert42);
	\draw[\myarrow, thick] (vert42) -- (vert44);
	\draw[\myarrow, thick] (vert44) -- (vert43);
	\draw[\myarrow, thick] (vert43) -- (vert41);
	\draw[\myarrow, thick] (vert45) -- (vert41);
	\draw[\myarrow, thick] (vert45) -- (vert43);


\end{tikzpicture}
  \caption{
    Examples of networks covered by the presented theory in \Cref{sec:theory}.
    The crossed-out network has no input and is consequently not a network by our definition.
  }
  \label{fig:app_network_examples}
\end{figure}
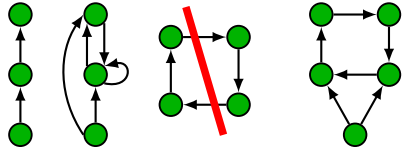

\subsection{Note: Connecting the theory of networks with deep neural networks}
\label{sec:theory_dnn}

For deep networks, nodes \(v\) correspond to layers and edges \(e\) to transformations between layers, such as convolutions (e.g., see networks in \Cref{fig:app_network_examples} and \Cref{fig:app_network_mnist}).
To a node \(v\) a state \(x_v \in \mathbb{R}^{D_v}\) is assigned, e.g.\ an image with \(D_v = 32 \times 32 \times 3\).
Let denote \(y_e\) the result of the transformation \(f_e\) which is specified by an edge \(e=(u,v)\):

\[
	y_e = f_e(\theta_e, x_{u}), 
\]

where \(\theta_e\) are parameters of the edge \(e\), e.g.\ a weight kernel.

For the node \(v\), let \(\text{SRC}_v\) denote the set of all edges targeting \(v\).
Ignoring the temporal dimension, a node's state is then computed as:

\[
	x_v = f_v(\vartheta_v, y_{e^1_v}, \ldots, y_{e_v^{|\text{SRC}_v|}}),
\]

where \(\vartheta_v\) are parameters of the vertex \(v\) (e.g., biases), and the mapping \(f_v\) specifying how the sources are combined (e.g., addition and/or multiplication).
Most architectures and network designs can be subsumed under this definition of a network, because we do not impose any constraints on the node and edge mappings \(f_v, f_e\).

For the experiments in this work, for every node \(v\) all results of incoming transformations were summed up:
\[
	x_v = f_v(b, y_{e^1_v}, \ldots, y_{e_v^{|\text{SRC}_v|}}) = \sigma\left( b + \sum\limits_{e \in \text{SRC}_v} y_{e} \right),
\]
where \(\sigma\) is some activation function, \(b\) is a channel-wise bias and the edge transformations \(f_{e}\) were convolutions with suitable stride to provide compatible dimensions for summation.

\subsection{Proofs for definition (rollout) in \Cref{def:rollout}}
\label{sec:proof_rollout}

Let \(N=(V,E)\) be a network.
We call a mapping \(R: E \rightarrow \{0, 1\}\) a \textbf{rollout pattern} of \(N\).
For a rollout pattern \(R\), the \textbf{rollout window} of size \(W\in\mathbb{N}\) is the directed graph \(R_W=(V_W, E_W)\) with:

\begin{equation}
\begin{split}
	V_W & \defeq \{0, \ldots, W\} \times V,\ \ \ \overline{v} = (i,v) \in V_W \\
	E_W & \defeq \{((i,u),(j,v)) \in V_W \times V_W \ \ | \ \ (u,v) \in E \ \land \ j = i + R((u,v))\}.
\end{split}
\end{equation}

We dropped the dependency of especially \(E_W\) on the rollout pattern \(R\) in the notation.
A rollout pattern and its rollout windows are called \textbf{valid} iff \(R_W\) is acyclic for one and hence for all \(W \in \mathbb{N}\).
We denote the set of all valid rollout patterns as \(\mathcal{R}_{N}\), and the rollout pattern for which \(R \equiv 1\) the \textbf{streaming rollout} \(R^\text{stream} \in \mathcal{R}_N\). We say two rollout patterns \(R\) and \(R^{\prime}\) are \textbf{equally model-parallel} iff for all edges \(e = (u,v) \in E\) not originating in the network's input \(u \notin I_N\) are equal \(R(e) = R^{\prime}(e)\).
For \(i \in \{0, \ldots, W\}\), the subset \(\{i\} \times V \subset V_W\) is called the \(i\)-th \textbf{frame}.

\paragraph{Note (interpretation):} Here, we show how this definition reflects the intuition, that a rollout should be consistent with the network in the sense that it should \textit{contain} all edges / nodes of the network and should not add new edges / nodes, which are not present in the network. Further, we show that this definition yields rollout windows which are temporally consistent and that rollout windows are consistent with regards to each other:
\begin{itemize}
	\item \textbf{Rollout windows cannot add \textit{new} edges / nodes:} By this, we mean, that a rollout window only contains derived nodes and edges from the original network and for example cannot introduce edges between nodes in the rollout window, which were not already present in the network. This follows directly from the definition of \(E_W\).
	\item \textbf{Edges / nodes of the network are contained in a rollout window:} For vertices this is trivial and for edges \(e = (u,v) \in E\) always \(((0,u),(R(e),v)) \in E_W\).
	\item \textbf{Rollout windows contain no temporal backward edges:} A backward edge is an edge \(((i,u),(j,v)) \in E_W\) with \(j < i\). But we know for all edges that \(j = i + R((u, v))\).
	\item \textbf{Temporal consistency:} Temporal consistency means that for an edge \(((i,u),(j,v)) \in E_W\) and a second edge between the \textit{same} nodes \(((i_{\star},u),(j_{\star},v)) \in E_W\) the temporal gap is the same \(j - i = j_{\star} - i_{\star}\). By definition, both are equal to \(R((u, v))\).
	\item \textbf{Rollout windows are compatible with each other:} We show that \(R_{W}\) is a sub-graph of \(R_{W + 1}\), in the sense that \(V_W \subset V_{W + 1}\) and \(E_{W} \subset E_{W + 1}\): From the definition, this is obvious for the set of vertices and edges, but nevertheless we will state it for edges anyway: Let \(\overline{e} \in E_W\) with \(\overline{e} = ((i,u),(j,v))\). Then by definition \((u,v) \in E\) and \(j = i + R((u,v))\). Hence, \(\overline{e} \in R_{W + 1}\).
\end{itemize}

\paragraph{Proof (definition of valid rollout pattern is well-defined):} For a rollout pattern, we prove that if the rollout window of a certain size \(W\) is valid, then the rollout window for any size is valid: Let \(R\) be a rollout pattern for a network \(N\) and \(R_W\) be a valid rollout window. Because \(R_W\) hence contains no cycles, also \(R_{W^{\prime}}\) for \(W^{\prime} < W\) contains no cycles (see statement about rollout window compatibility from above). Using induction, it is sufficient to show that \(R_{W + 1}\) is valid. Assuming it is not, let \(p\) be a cycle in \(R_{W + 1}\). Because there are no temporal backward edges (see above) \(p\) has to be contained in the last, the \((W + 1)\)-th frame. Because of the temporal consistency of rollout windows (see above), there are now cycles in all previous frames which contradicts the validity of \(R_W\).

\paragraph{Proof (streaming rollout exists and is valid):} The streaming rollout pattern \(R^\text{stream} \equiv 1\) always exists, because according to our network definition, \(E\) is not empty. Further, the streaming rollout pattern is always valid: Assuming that this is not the case, let \(R^{\text{stream}}_W\) be a rollout window of size \(W\) which is not acyclic and let \(p\) be a cycle in \(R^{\text{stream}}_W\). Because there are no backward edges \(\overline{e} = ((i,u),(j,v)) \in E_W\) with \(j < i\), all edges of the cycle must be inside a single frame, which is in contradiction to \(R^\text{stream} \equiv 1\).

\paragraph{Note (streaming rollout is un-ambiguous):} Considering the sets of all \textit{most streaming} and \textit{most non-streaming} rollout patterns
\[
	R_\text{streaming} = \left\{R \in \mathcal{R}_N \ \left| \ |R^{-1}(1)| = \max\limits_{R_{\star} \in \mathcal{R}_N} |R_{\star}^{-1}(1)| \right.\right\}
\]
\[
	R_\text{non-streaming} = \left\{R \in \mathcal{R}_N \ \left| \ |R^{-1}(0)| = \max\limits_{R_{\star} \in \mathcal{R}_N} |R_{\star}^{-1}(0)| \right.\right\}
\]
we have shown above that \(|R_\text{streaming}| = 1\) and this is exactly the streaming rollout. In contrast, \(|R_\text{non-streaming}| \geq 1\) especially for networks containing cycles with length greater \(1\). In this sense, the streaming rollout is un-ambiguous because it always uniquely exists while \textit{the} most-sequential rollout is ambiguous.

\subsection{Proof for Lemma 1 in \Cref{lemma1}}
\label{sec:proof_lemma1}

\paragraph{Lemma 1:} Let \(N=(V,E)\) be a network. 
The number of valid rollout patterns \(|\mathcal{R}_{N}|\) is bounded by:
\begin{equation}
	1 \leq n \leq |\mathcal{R}_{N}| \leq 2^{|E| - |E_{\text{rec}}|},
\end{equation}
where \(E_{\text{rec}}\) is the set of all self-connecting edges \(E_{\text{rec}} \defeq \{(u, v) \in E\ |\ u = v\}\), and \(n\) either:
\begin{itemize}
	{\item \(n = 2^{|E_\text{forward}|}\), with \(E_\text{forward}\) being the set of edges not contained in any cycle of \(N\), or
	\item \(n = \prod\limits_{p \in C}(2^{|p|} - 1)\), \(C \subset C_N\) being any set of minimal and pair-wise non-overlapping cycles.}
\end{itemize}

\paragraph{Proof \(|\mathcal{R}_{N}| \leq 2^{|E| - |E_{\text{rec}}|}\):} The number of all (valid and invalid) rollout patterns is \(2^{|E|}\), because the pattern can assign \(0\) or \(1\) to every edge. In order to be valid (acyclic rollout windows), the pattern has to assign \(1\) at least to every self-connecting edge.

\paragraph{Proof \(1 \leq n\):} Concerning the forward case: According to the definition of a network, \(I_N\) is not empty and hence there always exists at least one forward edge \(|E_\text{forward}| > 0\). Concerning the recurrent case: It is easy to see that \(n\) is greater than \(0\), increases with \(|C|\) and that \(C\) has to be at least the empty set.

\paragraph{Proof \(n \leq |\mathcal{R}_{N}|\) forward case:} Considering the streaming rollout pattern \(R^\text{stream} \equiv 1\) which always exists and is always valid (see above), we combinatorically can construct \(2^{|E_\text{forward}|}\) different valid rollout patterns on the basis of the streaming rollout pattern by combinatorically changing \(R(e)\) for all forward edges \(e \in E_\text{forward}\).

\paragraph{Proof \(n \leq |\mathcal{R}_{N}|\) recurrent case:} W.l.o.g. in case \(C_N = \emptyset\) we set \(n = 1\). Otherwise let \(C \subset C_N\) be any set of minimal and pair-wise non-overlapping cycles. Based on the streaming rollout pattern we will again construct the specified number of rollout patterns. The idea is that every cycle \(p \in C\) gives rise to \(2^{|p|} - 1\) different rollout patterns by varying the streaming rollout \(R^\text{stream}(E) \equiv 1\)
on all edges in \(p\) and we have to subtract the one rollout for which \(R(p) \equiv 0\), because for this specific rollout pattern, the cycle \(p\) does not get unrolled. Because the cycle is minimal, those \(2^{|p|} - 1\) patterns are different from one another. Because all cycles in \(C\) are disjunct we can combinatorically use this construction across all cycles of \(C\) and constructed \(\prod\limits_{p \in C} (2^{|p|} - 1)\) valid rollouts.

\subsection{Proof update steps convergence to full state in \Cref{theory:u_conv}}
\label{sec:proof_u_conv}

Let \(R_W\) be a rollout window for a valid rollout pattern \(R\) of the network \(N=(V,E)\). Then, starting from the initial state \(S_\text{init}\) and successively applying update steps \(U\), converges always to the full state \(S_\text{full}\): 
\[
\exists n \in \mathbb{N} \ :\ U^n(S_\text{init}) = S_\text{full}
\]

\paragraph{Proof:} Using induction, we show this without loss of generality for \(R_1\). Assuming that this is not the case, then there exists a state \(S \in \Sigma_1\), such that
\[
	\forall n \in \mathbb{N} \ : \ U^n(S) = S,\ \text{and}\ \exists \overline{v} = (1,v) \in V_1 \ :\ S(\overline{v}) = 0
\]
But being unable to update \(\overline{v}\) means, that there is another node that is input to \(\overline{v}\) which is also not updated yet \((1,v_1) \in V_1\) and \(S((1,v_1)) = 0\). Because there are no loops in \(R_1\) these nodes are not the same \(v \neq v_1\). This line of argument can now also be applied to \(v_1\) leading to a third node \((1,v_2)\) with \(S((1,v_2)) = 0\) and \(v \neq v_1 \neq v_2\) and so on. Because we only consider networks with \(|V| < \infty\) this leads to a contradiction.

\subsection{Proof Definition of inference tableau in \Cref{def:state_step_tableau}}
\label{sec:proof_inference_update_tableau}

For a valid rollout pattern \(R\) and a rollout window \(R_W\), we defined the inference tableau as the mapping \(T : V_W \rightarrow \mathbb{N}\) with:
\[
	T(\overline{v}) \ \defeq \ \max\limits_{p \in P_{\overline{v}}} |p| \ = \ \argmin_{n\in\mathbb{N}}\left\{U^{n}(S_\text{init})(\overline{v}) = 1 \right\} 
\]
For this, we have to show, that the equation holds.

\paragraph{Proof:}
We denote:
\[
T^\text{max}(\overline{v}) \defeq \max\limits_{p \in P_{\overline{v}}} |p| 
\]
\[
T^\text{min}(\overline{v}) \defeq \argmin_{n\in\mathbb{N}}\left\{U^{n}(S_\text{init})(\overline{v}) = 1 \right\}
\]
and have to show \(T^\text{min} \equiv T^\text{max}\). The proof is divided into two parts, first showing that the number of necessary update steps to update a certain node \(\overline{v}\) is higher or equal the length of any path \(p \in P_{\overline{v}}\) and hence \(T^\text{min} \geq T^\text{max}\). In the second part of the proof, we show that maximal paths \(p \in P_{\overline{v}}\) get successively updated at every update step.

In the first part, we will prove the following statement: For every \(\overline{v} \in V_W\) and \(p \in P_{\overline{v}}\):
\begin{equation}\label{eq:sublemma_inference_update_tableau}
	T^\text{min}(\overline{v}) \geq T^\text{min}(p(1)_\text{src}) + |p|.
\end{equation}
Here, we denoted again the edges of the path as \(p(i) = (p(i)_\text{src}, p(i)_\text{tgt}) \in E_W\).
In words this means, that for every path in a valid rollout window, the tableau values of the paths first \(p(1)_\text{src}\) and last \(\overline{v} = p(|p|)_\text{tgt}\) node differ at least about the length of the path. This is clear for paths of length one \(|p| = 1\), because \(p(1)_\text{tgt}\) can neither be updated before nor at the same update step as \(p(1)_\text{src}\), because \(p(1)_\text{src}\) is an input of \(p(1)_\text{tgt}\). Using induction and the same argument for paths of greater lengths \(|p|=n\) proves (\ref{eq:sublemma_inference_update_tableau}) and therefore also \(T^\text{min} \geq T^\text{max}\). 

In the second part of the proof, we will show that for all \(\overline{v} \in V_W\) all paths \(p \in P_{\overline{v}}\) of maximal length get updated node by node in each update step: 
\[
	U^{i-1}(S_\text{init})(p(i)_\text{tgt}) = 0
\]
\[
	U^{i}(S_\text{init})(p(i)_\text{tgt}) = 1
\]
for \(i \in \{1, \ldots, |p|\}\). 

We will prove this via induction over maximal path lengths. For \(\overline{v} \in V_W\) for which the maximum length of a path \(p \in P_{\overline{v}}\) is zero \(|p| = 0\) and hence \(P_{\overline{v}} = \emptyset\) we know by definition of \(P_{\overline{v}}\) and because the rollout window is connected to the initial state (see \Cref{sec:proof_u_conv}) that \(U^0(S_\text{init})(\overline{v}) = S_\text{init}(\overline{v}) = 1\). This proves the second part for \(\overline{v}\) with maximum path length zero. Now we consider \(\overline{v} \in V_W\) for which the maximum length of a path \(p \in P_{\overline{v}}\) is one \(|p| = 1\). Because \(p\) is maximal, its first node is in the initial state \(S_\text{init}(p(1)_\text{src}) = 1\) and due to the definition of \(P_{\overline{v}}\) it is \(S_\text{init}(p(1)_\text{tgt}) = 0\). Further, because \(p\) is maximal and of length \(1\), the initial state of all inputs to \(p(1)_\text{tgt}\) is \(1\) and hence \(p(1)_\text{tgt}\) can be updated in the first update step \(U(S_\text{init})(p(1)_\text{tgt}) = 1\). This proves the second part for \(\overline{v}\) with maximum path length one.

Let now be \(n \geq 2\), and we assume that the statement is true for nodes \(\overline{v}\) for which maximal paths \(p \in P_{\overline{v}}\) have length \(n\). Be \(\overline{v}\) now a node in \(V_W\) for which the maximal length of a path \(p \in P_{\overline{v}}\) is \(n + 1\). If the end node of a maximal path \(p \in P_{\overline{v}}\) cannot be updated \(U^{n+1}(S_\text{init})(p(n+1)_\text{tgt}) = 0\), then one of this end node's inputs \(\overline{v}_\text{input} \in V_W\) was not yet updated \(U^{n}(S_\text{init})(\overline{v}_\text{input}) = 0\). But because \(p\) is maximal and of length \(n+1\), and \(\overline{v}_\text{input}\) is input to \(\overline{v}\), the maximum length of paths in \(P_{\overline{v}_\text{input}}\) is \(n\). Hence \(U^{n}(S_\text{init})(\overline{v}_\text{input}) = 1\) contradicting that \(\overline{v}_\text{input}\) was not yet updated and therefore proving the second part of the proof. This proves \(T^\text{min} \equiv T^\text{max}\) and hence both can be used to define the inference tableau.

\subsection{Proof for Theorem 1 in \Cref{theorem:pip}}
\label{sec:proof_pip}

\paragraph{Theorem 1:}
Let \(R\) be a valid rollout pattern for the network \(N=(V,E)\) then the following statements are equivalent:
\begin{itemize}
{
	\item[a)] \(R\) and the streaming rollout pattern \(R^\text{stream}\) are equally model-parallel.
	\item[b)] The first frame is updated entirely after the first update step: \(F(R) = 1\).
	\item[c)] For \(W\in\mathbb{N}\), the \(i\)-th frame of \(R_W\) is updated at the \(i\)-th update step:
\[
  \forall (i,v) \in V_W \ : \  T((i,v)) \leq i.
\]
	\item[d)] For \(W\in\mathbb{N}\), the inference tableau of \(R_W\) is minimal everywhere and over all rollout patterns (most frequent responses \& earliest response):
\[
	\forall \overline{v} \in V_W \ :\ T_{R_W}(\overline{v}) = \min\limits_{R^{\prime} \in \mathcal{R}_N}T_{R^{\prime}_W}(\overline{v}).
\]
}\vspace*{-\baselineskip}
\end{itemize}

\paragraph{Proof:} 
Equivalency of statements a) - d) will be shown via a series of implications connecting all statements:

\paragraph{a) \(\implies\) b):} Assuming there is a \(\overline{v}=(1,v)\) which cannot be updated with the first update step, then there has to be an input \((1,v_\text{input})\) of \(\overline{v}\) for which \(S_\text{init}((1,v_\text{input})) = 0\) which contradicts that \(R\) is equally model-parallel to the streaming rollout.

\paragraph{b) \(\implies\) a):} Assuming \(R(e) = 0\) for an edge \(e = (u,v) \in E\) with \(u \notin I_N\), would yield a dependency of \((1,v)\) on \((1,u)\). Because \(u \notin I_N\), \((1,u)\) is not updated at the beginning \(S_\text{init}((1,u)) = 0\) and therefore \(U^{1}(S_\text{init})((1,v)) = 0\) and hence \(T((1,v)) \geq 2\) which contradicts b).

\paragraph{c) \(\implies\) b):} Trivial.

\paragraph{a) \(\implies\) c):} Let \(\overline{v} = (i,v) \in V_W\). First we note, that every maximal path \(p \in P_{\overline{v}}\) has to start in the initial state \(S_\text{init}(p(1)_\text{src}) = 1\), otherwise we can extend \(p\) to a longer path. We will use the definition of \(T\) over maximum path lengths to prove c). Let \(R\) be equally model-parallel to the streaming rollout and \(p \in P_{\overline{v}}\) a path of maximal length. We know \(S_\text{init}(p(1)_\text{src}) = 1\) and hence either \(p(1)_\text{src} \in \{0\} \times V\) or \(p(1)_\text{src} \in \{0,\ldots,W\} \times I_N\). For the first case, it is easy to see that \(|p| = i\), because \(R\) is equally model-parallel to the streaming rollout and hence one frame is bridged \(R(e) = 1\) for every edge \(e\) in \(p\). For the second case \(p(1)_\text{src} \in \{0,\ldots,W\} \times I_N\), it follows from the same argument as before that \(|p| = i - i_\text{src}\) with \(p(1)_\text{src} = (i_\text{src}, v_\text{src})\) which proves c).

\paragraph{a) \(\implies\) d):} 
For this proof we introduce \textbf{induced paths}: Let \(R\) be a valid rollout pattern, \(\overline{v} = (i,v) \in R_W\) and \(p_R \in P^{R_W}_{\overline{v}}\) (same as \(P_{\overline{v}}\) from rollout definition but now expressing the dependency on the rollout window \(R_W\)):
\begin{equation*}
\begin{split}
	p_R(k)  & = \overline{e}^{k} \\
			& = \left((j^{k}_\text{src}, e^{k}_\text{src}), (j^{k}_\text{tgt}, e^{k}_\text{tgt})\right) \\
			& = \left((j^{k}_\text{src}, e^{k}_\text{src}), (j^{k}_\text{src} + R(e^{k}), e^{k}_\text{tgt})\right),
\end{split}
\end{equation*}
for \(k \in \{1,\ldots,|p_R|\}\) and \(e^{k} = (e^{k}_\text{src}, e^{k}_\text{tgt}) \in E\). 
Let \(R^{\prime}\) be a second valid rollout pattern and let denote \(n = |p_R|\).
Notice that \((j^{n}_\text{tgt}, e^{n}_\text{tgt}) = (i, v)\).
We want to define the induced path \(p_{R^{\prime}} \in P^{R^{\prime}_W}_{\overline{v}}\) as the path also ending at \(\overline{v} \in R^{\prime}_W\), backwards using the \textit{same} edges as  \(p_R\) and respecting the rollout pattern \(R^{\prime}\).
We define this induced path \(p_{R^{\prime}} \in P^{R^{\prime}_W}_{\overline{v}}\) of \(p_{R}\) recursively, beginning with the last edge of \(p_R\), as the end of the following sequence of paths, starting with the path:
\begin{equation*}
\begin{split}
	p_{R^{\prime}, 1}   & : \{1\} \rightarrow E_{R^{\prime}_W} \\
						& p_{R^{\prime}, 1}(1) = ((i - R^{\prime}(e^{n}), e^{n}_\text{src}), (i, e^{n}_\text{tgt}))
\end{split}
\end{equation*}
Recursively we define:
\begin{equation*}
\begin{split}
	p_{R^{\prime}, m}   & : \{1,\ldots,m\} \rightarrow E_{R^{\prime}_W} \\
						& p_{R^{\prime}, m}(k) = p_{R^{\prime}, m - 1}(k - 1),\ k \in \{2,\ldots,m\} \\
						& p_{R^{\prime}, m}(1) = ((i - s_{R^{\prime},p_R}(m), v^{n - m + 1}_\text{src}), (i - s_{R^{\prime},p_R}(m - 1), v^{n - m + 1}_\text{tgt}))
\end{split}
\end{equation*}
with \(s_{R^{\prime},p_R}(m) = \sum\limits_{k=1}^{m} R^{\prime}(e^{n - k + 1})\). In words, \(s_{R^{\prime},p_R}(m)\) is the \textit{frame length} of the last \(m\) edges of the path \(p_R\) under the rollout pattern \(R^{\prime}\). The sequence stops at a certain \(m\), either if no edges are left in \(p_R\): \(m = n\) or at the first time the source of the path's first edge reaches the \(0\)-th frame: \(i - s_{R^{\prime},p_R}(m) = 0\). With this definition we can proceed in the prove of a) \(\implies\) d):

Let \(R\) be equally model-parallel to the streaming rollout pattern, \(W \in \mathbb{N}\), and \(\overline{v} \in V_W\). 
Let further be \(p_{R} \in P^{R_W}_{\overline{v}}\) a path of maximal length, \(R^{\prime}\) be any valid rollout pattern, and \(p_{R^{\prime}}\) be the induced path of \(p_{R}\). We want to show that \(|p_{R}| = |p_{R^{\prime}}|\).

If both rollouts are equally model-parallel on the edges of the path \(\{e^{1},\ldots,e^{|p_R|}\}\) (this means \(R(e^{k}) = R^{\prime}(e^{k})\) for \(k \in \{1,\ldots,|p_R|\}\) if \(e^{1}\) does not originate in the input \(e^{1}_\text{src} \notin I_N\), and for \(k \in \{2,\ldots,|p_R|\}\) if \(e^1\) does originate in the input), the path \(p_R\) and its induced path \(p_{R^{\prime}}\) are the same up to their first edge which might or might not bridge a frame, but in both cases \(|p_{R}| = |p_{R^{\prime}}|\). 

If the rollouts are not model-parallel on the edges of the path and hence differ on at least one edge \(e^k\) which does not originate in the input, and because \(R\) is equally model-parallel to the streaming rollout, it is: 
\begin{equation} \label{ad:sub1}
	s_{R,p_R}(|p_R|) > s_{R^{\prime},p_R}(|p_{R^{\prime}}|).
\end{equation}
Because the induced path using the same rollout cannot loose length, we also know:
\begin{equation} \label{ad:sub2}
	i - s_{R,p_R}(|p_R|) \geq 0.
\end{equation}
Greater than zero would be the case for \(p_R\) originating in the input \(p_R(1)_\text{src} \in \{1,\ldots,W\} \times I_N\). 
Combining (\ref{ad:sub1}) and (\ref{ad:sub2}) yields:
\begin{equation*}
	i - s_{R^{\prime},p_R}(|p_{R^{\prime}}|) > 0.
\end{equation*}
Considering the two stopping criteria from the sequence of paths used to define the induced path from above, this proves \(|p_{R}| = |p_{R^{\prime}}|\).

We now have proven that the induced path \(p_{R^{\prime}}\) from a maximal path \(p_{R}\) in a rollout window from a rollout pattern \(R\) which is equally model-parallel to the streaming rollout is never shorter than \(p_R\) (especially for highly sequential \(R^{\prime}\), most \(p_{R^{\prime}}\) are not of maximal length). This means, that the maximal length of paths in \(P^{R^{\prime}_W}_{\overline{v}}\) is at least as large as the maximal length of paths in \(P^{R_W}_{\overline{v}}\) which by definition of the inference tableau proves a) \(\implies\) d).

\paragraph{d) \(\implies\) b):} Trivial.

\section{Details about networks, data, and training}
\paragraph{}\label{sec:details_nets}

\begin{figure}[H]
  \centering
  \input{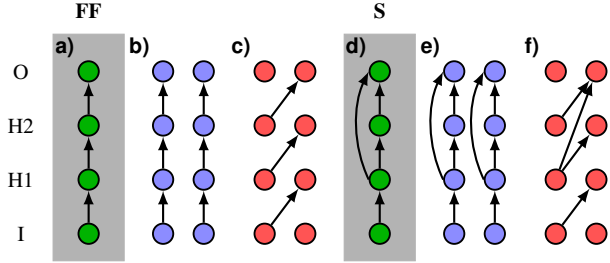}
  \caption{
    Neural networks (gray boxes) used for MNIST (\Cref{fig:results}a-c) with different rollouts.
    Schematics of a feed-forward network (FF, \textbf{a}, green) with its corresponding sequential (\textbf{b}, blue) and streaming (\textbf{c}, red) rollouts.
    Nodes represent layers, edges represent transformations, e.g.\ convolutions.
    Only one rollout step is shown and each column in (b) and (c) is one frame within the rollout.
    Rollouts are also shown for networks with an additional skip connection (S, \textbf{d-f}).
    Node labels on the left are referred to in \Cref{sec:details_nets}.
  }
  \label{fig:app_network_mnist}
\end{figure}

\begin{figure}[H]
  \centering
  \input{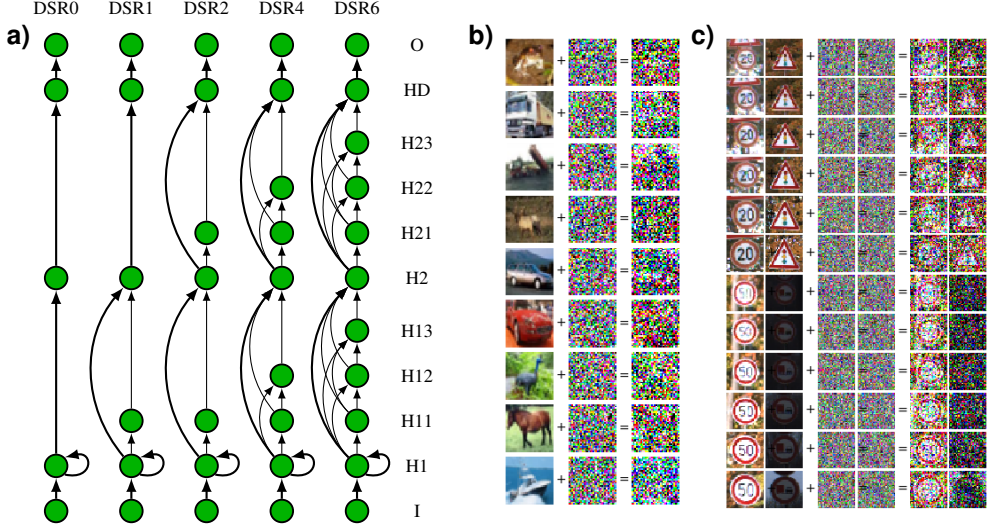}
  \caption{
    \textbf{a:} A selection of the sequence of networks evaluated on CIFAR10 (for details see \Cref{sec:details_nets}).
    For evaluating the GTSRB dataset the network DSR4 is used, but without the self-connection of node H1.
    The input of the networks are images with added Gaussian noise as shown in \textbf{b:} for CIFAR10 and \textbf{c:} for GTSRB (for details see \Cref{tab:datasets_setups}).
  }
  \label{fig:methods}
\end{figure}

In the depiction of network architectures (\Cref{fig:intro}, \Cref{fig:methods}, and  \Cref{fig:app_network_mnist}), connections between nodes are always realized as convolutional or fully connected layers.
In case a node (layer) is the target of several connections, its activation is always computed as the sum over outputs of these connections.
This is mathematically equivalent to concatenating all inputs of the layer and applying a single convolution on the concatenation.

\paragraph{MNIST}

The network designs are shown in \Cref{fig:intro} and \Cref{fig:app_network_mnist}. The size of the layers (pixels, pixels, features) are: input image I with (28, 28, 1), hidden layer H1 with (7, 7, 16), hidden layer H2 with (1, 1, 128) and output layer O with (1, 1, 10).

The following network design specifications were applied with A-B meaning the edge between layer A and layer B.
Some of these edges only exist in the networks with skip connection (S) or with skip and self-recurrent connections (SR).
For node labels see \Cref{fig:app_network_mnist}: 
\begin{itemize}
	\item I-H1: a convolution with receptive field \(7\) and stride \(4\)
	\item H1-H2 and H2-O: fully connected layers
	\item H1-O: a fully connected layer
	\item H1-H1-recurrence: a convolution with receptive field \(3\) and stride \(1\)
\end{itemize}

\paragraph{CIFAR10}
The network design is shown in \Cref{fig:methods}a.
We used a sequence of \(7\) increasingly deep network architectures with the first network DSR0 being a simple (\(3\) hidden layers) forward design and the first hidden layer having a self-recurrent connection.
We added additional hidden layers to generate the next networks in the following way: H11 to DSR0, H21 to DSR2, H12 to DSR3, ..., H23 to DSR6.
Note that every network is a sub-network of its successor. Hence, the length of the shortest path is always \(4\), while the length of the longest path increases from \(4\) to \(11\) by \(1\) for every consecutive network.

The size of the layers (pixels, pixels, features) are: input image I with \((32,32,3)\) and hidden layers H1, H11, H12, H13 with \((32,32,32)\) and H2, H21, H22, H23 with \((16,16,64)\), fully connected layer HD with \((4,4,512)\) and output layer O with \((1,1,10)\).

The following network design specifications were applied:

\begin{itemize}
	\item I-H1: a convolution with receptive field \(5\) and stride \(1\)
	\item H1-H11, H11-H12, H12-H13: a convolution with receptive field \(3\) and stride \(1\)
	\item H2-H21, H21-H22, H22-H23: a convolution with receptive field \(3\) and stride \(1\)
	\item H13-H2: convolutions with receptive field \(3\) and stride \(2\)
	\item H23-HD: convolutions with receptive field \(3\) and stride \(4\)
	\item H1-H1-recurrence: a convolution with receptive field \(3\) and stride \(1\)
	\item skip connections H1-H12, H1-H13, H1-H2, H11-H13, H11-H2, H12-H2 and H2-H22, H2-H23, H2-HD, H21-H23, H21-HD, H22-HD: convolution with receptive field \(3\) and stride \( \frac{\text{input size}}{\text{output size}} \)
\end{itemize}

\paragraph{GTSRB}
For the experiments, the network DSR4 shown in \Cref{fig:methods}a was used without the self-recurrence H1-H1 connection. Design specifications are adapted from the CIFAR10 networks with input image I with \((32,32,3)\) and output layer O with \((1,1,43)\). 
For each repetition, \(80\%\) of the data was randomly taken for training, \(10\%\) for validation, and \(10\%\) for testing.

\paragraph{Training details}
To train networks, we used RMSprop (\citep{tieleman2012rmsprop}) with an initial learning rate of \(10^{-4}\) and an exponential decay of \(10^{-6}\).
All networks were trained for \(100\) epochs.
A dropout rate of \(0.25\) was used for all but the last hidden layer, for which a rate of \(0.5\) was used. 
The loss for the rolled-out networks is always the mean over the single-frame prediction losses, for which we used cross-entropy.
At the zero-th frame, states of all but the input layers were initialized with zero.

Details about experimental setups and data processing are given in \Cref{tab:datasets_setups}.

\begin{table}[H]
\centering
\begin{small}
\begin{tabular}{lllllll}
\toprule
Data & value & perturbation & augmentation & training / & batch & reps \\
 & range &  &  & val. / test size & size &  \\
\midrule 
Noisy & [0,1] & 1. $\mathcal{N} (\sigma=2.0)$; & None & 50k / 10k / 10k & 128 & 6 \\
MNIST &  & 2. clipped to [0,1] &  &  &  &  \\
\addlinespace
CIFAR10 & [0,1] & 1. $\mathcal{N} (\sigma=1.0)$; & horizontal & 40k / 10k / 10k & 64 & 1 \\
 &  & 2. clipped to [0,1] & flipping &  &  &  \\
 &  & 3. mean subtracted &  &  &  &  \\
\addlinespace
GTRSB & [0,1] & 1. $\mathcal{N} (\sigma=0.5)$; & None & 80\% / 10\% / 10\% of & 16 & 12 \\
 &  & 2. clipped to [0,1] &  & 1305 tracks &  &  \\
 &  & 3. resized to  $32 \times 32$ pixels &  & (30 frames each) &  & \\ 
\bottomrule
\end{tabular}
\end{small}
\caption{Experimental setups for the data sets: Image pixels were scaled (\textit{value range}); then each frame was perturbed adding Gaussian noise with a standard deviation of $\sigma$, clipped back into the value range and for CIFAR10 the channel-wise mean over all training images was subtracted. For GTRSB images of different size were resized. Data \textit{augmentation} was conducted for training and the number of images for training, validation, and testing (\textit{training / val. / test size}) and the \textit{batch sizes} are listed. Experiments were repeated (\textit{reps}) times.}
\label{tab:datasets_setups}
\end{table}

\section{Toolbox for streaming rollouts}
\paragraph{}\label{sec:statestream_toolbox}

\begin{figure}[H]
\begin{center}
\includegraphics[width=0.75\textwidth]{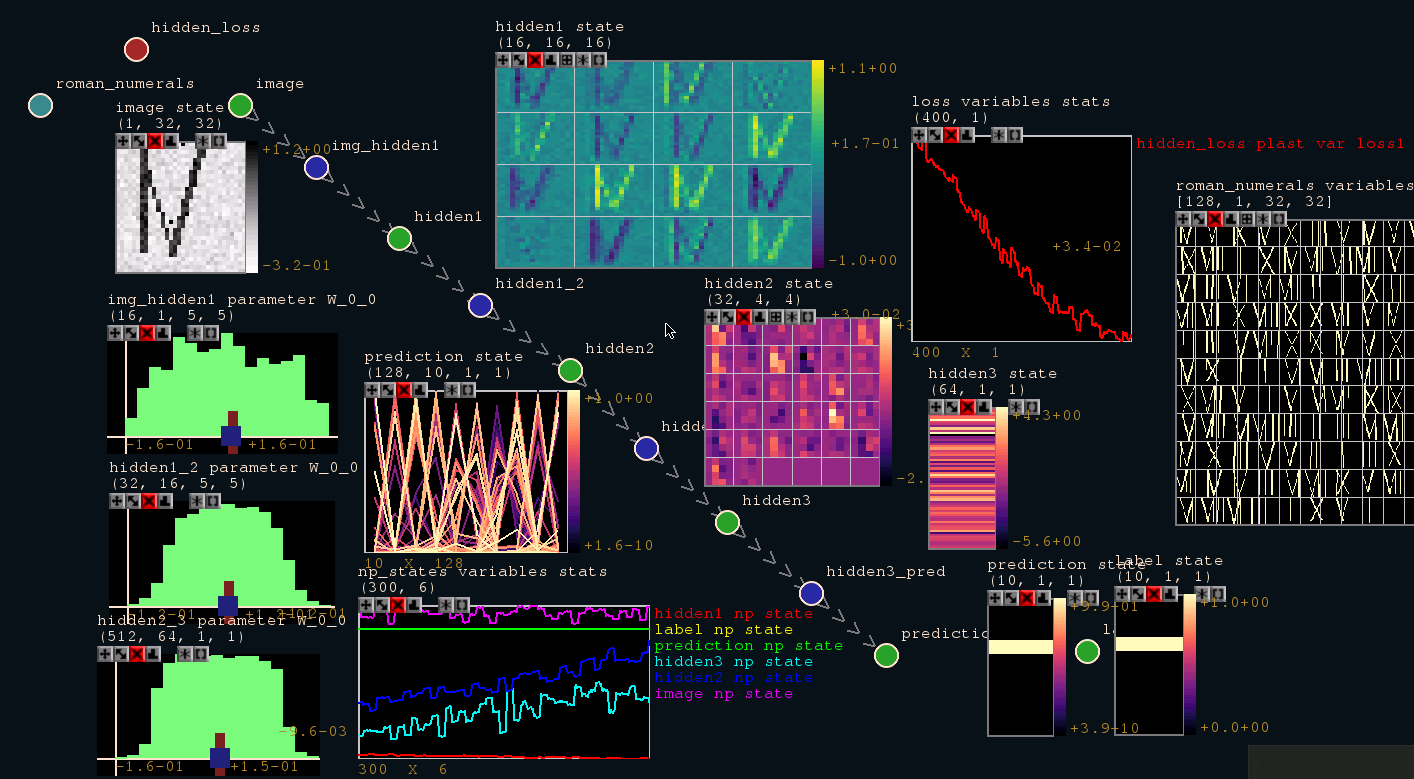}
\end{center}
\caption{Visualization example of a simple classification network using the provided toolbox (best viewed in color). The network is shown as graph together with information about the network.}
\label{fig:figure_statestream}
\end{figure}

One of the contributions of this work is to provide an open-source toolbox (\url{https://github.com/boschresearch/statestream}) to design, train, evaluate, and interact with the streaming rollout of deep networks. 
An example screenshot of the provided user interface is shown in~\Cref{fig:figure_statestream}.

Networks are specified in a text file, and a core process distributes the network elements onto separate processes on CPUs and/or GPUs. 
Network elements are executed with alternating read and write phases, synchronized via a core process, and operate on a shared representation of the network. The toolbox is written in Python and uses the Theano \citep{2016arXiv160502688short} or TensorFlow \citep{tensorflow2015_whitepaper} backend. The shared representation enables parallelization of operations across multiple processes and GPUs on a single machine and enables online interaction.

\end{document}